\def\year{2020}\relax
\documentclass[letterpaper]{article} 
\usepackage{aaai20}  
\usepackage{times}  
\usepackage{helvet} 
\usepackage{courier}  
\usepackage[hyphens]{url}  
\usepackage{graphicx} 
\usepackage{graphicx}  
\usepackage{subfigure}
\usepackage{amsmath}
\usepackage{amssymb}
\usepackage{multirow}
\usepackage{array}
\usepackage{booktabs}
\usepackage{algorithm}
\usepackage{algorithmic}

\title{Planar Prior Assisted PatchMatch Multi-View Stereo}
\author{Qingshan Xu and Wenbing Tao
	\thanks{Corresponding author}\\ 
	National Key Laboratory of Science and Technology on Multispectral Information Processing\\ 
	School of Artifical Intelligence and Automation, Huazhong University of Science and Technology, China\\
	\{qingshanxu, wenbingtao\}@hust.edu.cn 
}
 
\begin{document}

\maketitle

\begin{abstract}
The completeness of 3D models is still a challenging problem in multi-view stereo (MVS) due to the unreliable photometric consistency in low-textured areas. Since low-textured areas usually exhibit strong planarity, planar models are advantageous to the depth estimation of low-textured areas. On the other hand, PatchMatch multi-view stereo is very efficient for its sampling and propagation scheme. By taking advantage of planar models and PatchMatch multi-view stereo, we propose a planar prior assisted PatchMatch multi-view stereo framework in this paper. In detail, we utilize a probabilistic graphical model to embed planar models into PatchMatch multi-view stereo and contribute a novel multi-view aggregated matching cost. This novel cost takes both photometric consistency and planar compatibility into consideration, making it suited for the depth estimation of both non-planar and planar regions. Experimental results demonstrate that our method can efficiently recover the depth information of extremely low-textured areas, thus obtaining high complete 3D models and achieving state-of-the-art performance.       
\end{abstract}

\section{Introduction}

Multi-view stereo (MVS) aims to estimate the dense 3D model of the scene from a given set of calibrated images. Due to its wide applications in virtual/augmented reality and 3D printing and so on, much progress has been made in this domain~\cite{Furukawa2010Accurate,Strecha2006Combined,Merrell2007Real,Goesele2007Multi,Liu2009Continuous,Schonberger2016Pixelwise} in the last few years. However, recovering a dense and realist 3D model is still a challenging problem since the depth estimation in low-textured areas always fails.

\begin{figure}[t]
	\centering
	\subfigure[Reference image]{
		\includegraphics[width=0.45\linewidth]{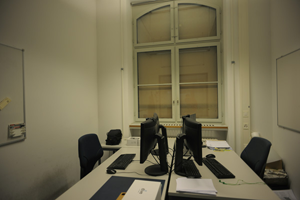}}
	\subfigure[Ground truth]{
		\includegraphics[width=0.45\linewidth]{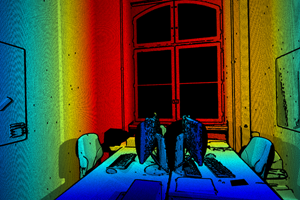}}
	\subfigure[Triangulation]{
		\label{fig:subfig:motic}
		\includegraphics[width=0.45\linewidth]{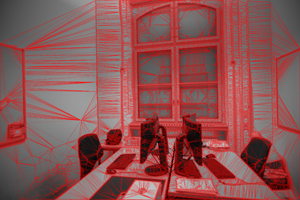}}
	\subfigure[ACMP (Ours)]{
		\label{fig:subfig:motid}
		\includegraphics[width=0.45\linewidth]{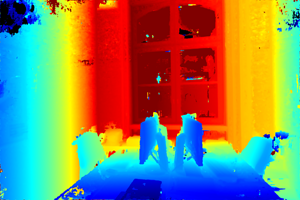}}
	\caption{Illustration of discrimination and depth maps obtained by our method. The piecewise planar priors (c) through triangulation can adaptively gain discrimination for different low-textured areas. This helps to obtain better depth estimation for large low-textured areas (d).}
	\label{fig:motivation}
\end{figure}

The failure of depth estimation in low-textured areas mainly comes from the unreliable photometric consistency measure in these areas. As low-textured areas always appear in smooth homogeneous surfaces (Figure~\ref{fig:motivation}), many methods~\cite{Woodford2009Global,Gallup2010Piecewise} assume that these surfaces are piecewise planar. Then, they formulate this prior as a regularization term in a global energy framework to recover the depth estimation in low-textured areas. Due to the difficulty in solving such optimization problems, the efficiency of these methods is low and they are easy to be trapped in local optima. Recently, PatchMatch multi-view stereo methods~\cite{Zheng2014PatchMatch,Galliani2015Massively,Schonberger2016Pixelwise,Xu2018Multi} become popular as their used PatchMatch-based optimization~\cite{Barnes2009PatchMatch} makes depth map estimation efficient and accurate. As these methods do not explicitly model the planar priors, these methods still encounter the failure in low-textured areas. Based on the individual advantages of planar prior models and PatchMatch multi-view stereo, we expect to construct a planar prior assisted PatchMatch multi-view stereo framework to efficiently recover the depth estimation in low-textured areas.  

To embed the planar prior models into PatchMatch multi-view stereo, in this work we rethink the right way to build the multi-view aggregated matching cost and propose a planar prior assisted PatchMatch multi-view stereo framework to help the depth estimation in low-textured areas. In conventional PatchMatch multi-view stereo methods,  sparse credible correspondences can be distinguished in discriminative regions, such as edges and corners. As these correspondences always coincide with the vertices in the mesh representation of 3D models, this means the sparse credible correspondences almost constitute the skeleton of a 3D model. Therefore, we first triangulate these correspondences to produce planar models. Note that, the planar priors are also suited for non-planar regions as credible correspondences are very dense in these regions and can adaptively form triangular primitives of different sizes. To derive the planar prior assisted multi-view aggregated matching cost, we leverage a probabilistic graphical model to simultaneously model photometric consistency and planar compatibility. The planar compatibility constrains predicted depth estimates to fall within an appropriate depth range while the photometric consistency can better reflect the depth changes in well-textured areas. At last, to alleviate the influence of unreliable planar priors, multi-view geometric consistency is enforced to rectify erroneous depth estimates.     

In a nutshell, our contributions are as follows: {\bf 1)} We propose a novel planar prior assisted PatchMatch multi-view stereo framework for multi-view depth map estimation. This framework not only inherits the high efficiency of PatchMatch multi-view stereo but also leverages planar priors to help the depth estimation in low-textured areas. {\bf 2)} We adopt a probabilistic graphical model to induce a novel multi-view aggregated matching cost. This novel cost function takes both photometric consistency and planar compatibility into consideration. We demonstrate the effectiveness of our method by yielding state-of-the-art dense 3D reconstructions on ETH3D benchmark~\cite{Schops2017Multi}. Our code will be available at \url{https://github.com/GhiXu/ACMP}.

\section{Related Work}

Our work is relevant to both PatchMatch multi-view stereo and planar priors, therefore we will review relevant literature in these areas. 

\noindent\textbf{PatchMatch Multi-View Stereo} PatchMatch multi-view stereo methods exploit the core idea of PatchMatch~\cite{Barnes2009PatchMatch}, sampling and propagation , to effectively estimate depth maps for each image. Focusing on different problems, many PatchMatch multi-view stereo methods have been proposed. \cite{Zheng2014PatchMatch,Schonberger2016Pixelwise} jointly estimate depth maps and pixelwise view selection by a probabilistic graphical model. \cite{Galliani2015Massively} utilizes a diffusion-like propagation scheme to make better use of the parallelization of GPUs. By inheriting the checkerboard pattern of \cite{Galliani2015Massively}, ACMH~\cite{Xu2018Multi} designs an adaptive checkerboard sampling strategy to propagate more reliable hypotheses. Moreover, ACMH further exploits these hypotheses to infer pixelwise view selection. However, as the photometric consistency on which these methods depend cannot get reliable discrimination in low-textured areas, the depth estimation of these methods always fails in these areas. To get reliable discrimination from low-textured areas, \cite{Wei2014Multi} leverages multi-scale scheme to achieve this at low resolution images. Then, it propagates the discrimination to the original resolution images by considering the relative depth difference from all neighboring views. On the multi-scale scheme, ACMM~\cite{Xu2018Multi} further considers the influence of view selection and leverages multi-scale geometric consistency to propagate the discrimination. Additionally, \cite{Romanoni2019TAPA} extracts superpixels at two scales and constrains the hypotheses in low-textured areas by the planes fitted for each superpixel. However, the discrimination obtained by the multi-scale scheme sometimes is limited by the predefined scales especially for large low-textured areas. In contrast, we leverage piecewise planar priors built from triangulation to adaptively acquire the discrimination for different low-textured areas.

\noindent\textbf{Planar Priors} Planar prior models are popular in 3D reconstruction as many scenes can be represented by a variety of plane primitives, especially for man-made environments. \cite{Gennert1988Brightness,Woodford2009Global} formulate planar prior models as second-order smoothness priors in a global energy function framework. This leads to a triple clique representation in the global energy function, making the optimization very difficult. To distinguish planar regions and non-planar objects in urban scenes, \cite{Gallup2010Piecewise} train a planar classifier to obtain the raw segmentation results and combine these segments with multi-view photometric consistency to define a global energy function to refine the predictions. Besides, it is worth noting that there exist some methods~\cite{Geiger2010Efficient,Zhang2015MeshStereo} employing triangulation to construct planar priors for disparity estimation in stereo matching. \cite{Geiger2010Efficient} forms a triangulation on robustly matched correspondences to build a prior over the disparity space. This can not only reduce the disparity search space, but also recover low-textured surfaces. To simultaneously output a disparity map and a 3D triangulation mesh, \cite{Zhang2015MeshStereo} first partitions stereo images into 2D triangles with shared vertices. Then, it formulates a two-layer Markov random field to jointly model disparity maps and vertex splitting probabilities. This assumes that the scene structure is piecewise planar and imposes regularization in the region-based stereo. Similar to \cite{Geiger2010Efficient}, our method also leverages the triangulation to build piecewise planar priors. Differently, embedding the priors into multi-view stereo is nontrivial as it needs to consider the visibility and geometry constraints of different views. To this aim, we embrace planar priors with PatchMatch MVS to consider the visibility and geometric constraints of different views.

\begin{figure*}[t]
	\centering
	\includegraphics[width=0.137\textwidth]{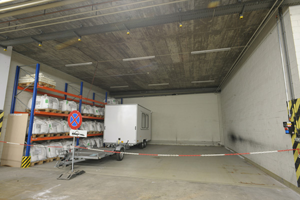} 
	\includegraphics[width=0.137\textwidth]{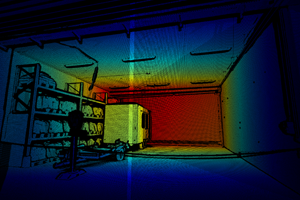} 
	\includegraphics[width=0.137\textwidth]{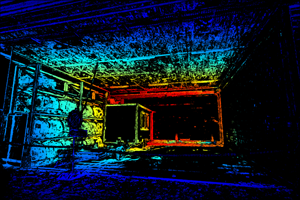}
	\includegraphics[width=0.137\textwidth]{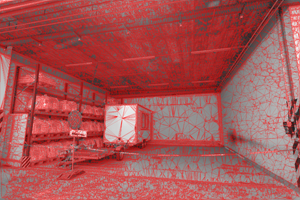}
	\includegraphics[width=0.137\textwidth]{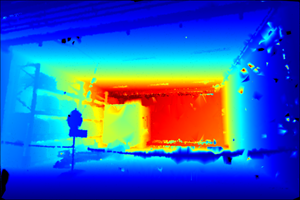}
	\includegraphics[width=0.137\textwidth]{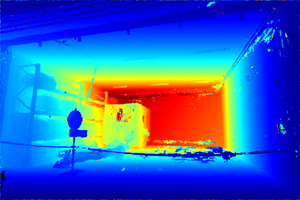}
	\includegraphics[width=0.137\textwidth]{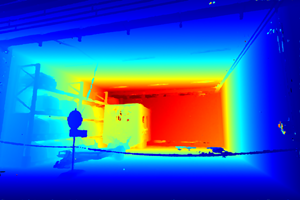}
	\includegraphics[width=0.137\textwidth]{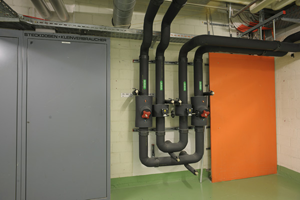} 
	\includegraphics[width=0.137\textwidth]{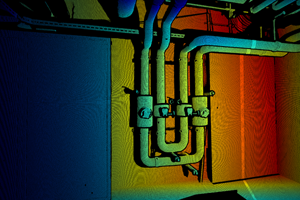} 
	\includegraphics[width=0.137\textwidth]{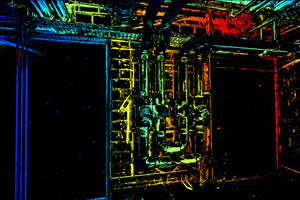}
	\includegraphics[width=0.137\textwidth]{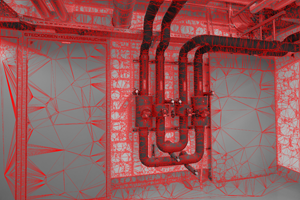}
	\includegraphics[width=0.137\textwidth]{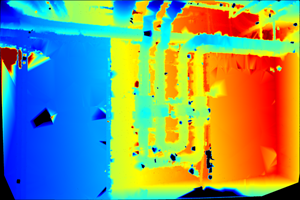}
	\includegraphics[width=0.137\textwidth]{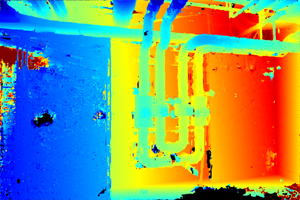}
	\includegraphics[width=0.137\textwidth]{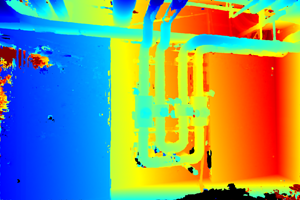}
	\begin{tabular}{p{0.116\textwidth}<{\centering}p{0.116\textwidth}<{\centering}p{0.116\textwidth}<{\centering}p{0.116\textwidth}<{\centering}p{0.118\textwidth}<{\centering}p{0.118\textwidth}<{\centering}p{0.116\textwidth}<{\centering}}
		(a) & (b) & (c) & (d) & (e) & (f) & (g) \\
	\end{tabular}
	\caption{(a) images; (b) ground truth; (c) sparse correspondences; (d) triangulation; (e) planar model directly calculated from (d); (f) planar prior assisted PatchMatch MVS; (g) geometric consistency.}
	\label{fig:Pipeline}
\end{figure*}

\section{Planar Prior Assisted PatchMatch MVS}  

Suppose we have a set of input images $\mathcal{I}=\{\emph{I}_{m}\,|\,m=1{\cdots}N\}$ with their corresponding camera parameters $\mathcal{P}=\{\emph{P}_{m}\,|\,m=1{\cdots}N\}$. Each image will be sequentially taken as a reference image $I_{\text{ref}}$ while the other images are source images $\mathcal{I}_{\text{src}}=\{\emph{I}_{j}\,|\,I_{j}\in\mathcal{I}\wedge I_{j}\neq I_{\text{ref}}\}$. Our work focuses on estimating the depth map of $I_{\text{ref}}$ in turn.

Currently, there exist two popular PatchMatch multi-view stereo frameworks, including sequential propagation pattern~\cite{Bailer2012Scale,Schonberger2016Pixelwise} and checkerboard propagation pattern~\cite{Galliani2015Massively,Xu2018Multi}. As pointed out in \cite{Xu2018Multi}, the latter one is more efficient and effective than the former one, thus we will build our algorithm on the checkerboard propagation pattern. 

Different from the conventional PatchMatch multi-view stereo methods, our method also takes as input the sparse credible correspondences of the reference image . To estimate the depth information of low-textured areas, our method consists of two stage. In the first stage, sparse correspondences are generated by conventional PatchMatch MVS methods and thresholding. Then, we triangulate these correspondences to produce planar models. In the second stage, we jointly consider the previous obtained planar models and photometric consistency by constructing a probabilistic graphical model. This derives a novel multi-view aggregated matching cost. By embedding this novel cost to the pipeline of PatchMatch MVS, we can obtain good depth estimation for low-textured areas.  

\subsection{Planar Model Construction} 

To start our algorithm, we implement the method of \cite{Xu2018Multi} to obtain sparse credible correspondences. The method follows the pipeline of PatchMatch MVS, iteratively performing adaptive checkerboard sampling and propagation, hypothesis updating via multi-view aggregated photometric consistency cost and refinement. A depth estimate will be considered as a credible correspondence if its final cost is lower than $0.1$ (Figure~\ref{fig:Pipeline}c).  

Given the sparse credible correspondences of $I_{\text{ref}}$, they always characterize the structure of a scene. Although the depth estimation for low-textured areas is lost, people can imagine the whole 3D model of a scene according to these correspondences. Based on this observation, we first triangulate these sparse credible correspondences to adaptively generate triangular primitives of different sizes. As can be seen from Figure~\ref{fig:Pipeline}d, the triangular primitives in well-textured areas are relatively small so that the structures of non-planar regions can be kept. On the other hand, the triangular primitives in low-textured areas are as large as possible to incorporate the information of credible correspondences.  

For each triangular primitive, we use its corresponding three vertices to calculate its plane parameters in the coordinate of the reference camera, including depth information and normal information. The pixels inside the same triangular primitive share the same plane parameters. Figure~\ref{fig:Pipeline}e shows two examples of planar models. It can be observed that the priori plane parameters can almost coincide with the optimal estimates for low-textured areas. It is worth noting that the structures of thin objects are also described by the generated triangular primitives. 

\subsection{Planar Prior Assistance}

With the priori plane hypotheses, the depth estimate for low-textured areas can be better approximated. However, these priori plane hypotheses also lead to many blocking artifacts in well-textured areas, especially in boundaries, whose depth information should be estimated well by photometric consistency. To take both photometric consistency and piecewise planar priors into consideration, we employ a probabilistic model to achieve this.

\noindent\textbf{Random Initialization} As a first step, we randomly generate a plane hypothesis $\theta_{l}=[d_{l},\boldsymbol{n}_{l}]$ for each pixel ${l}$ in the reference image, where $d_{l}$ is distance from a 3D plane to the origin and $\boldsymbol{n}_{l}$ is normal vector. We first calculate a matching cost for each source image via the plane hypothesis induced homography~\cite{Hartley2004Multiple}. Then, the initial multi-view aggregated matching cost for each hypothesis is computed by averaging the top-$K$ smallest matching costs.

\noindent\textbf{Hypothesis Sampling and Propagation} Following \cite{Galliani2015Massively,Xu2018Multi}, we divide all pixels in the reference image into a Red-Black pattern. This allows to use the hypotheses of red pixels to update those of black pixels and vice versa. Then, we use the adaptive checkerboard sampling of \cite{Xu2018Multi} to propagate eight neighboring good hypotheses to the current pixel to be estimated. These propagated hypotheses together with the current hypothesis $\theta_{0}$ of the pixel constitute the current candidate hypothesis set, $\boldsymbol{\theta}=\{\theta_{i}\,|\,i=0\cdots8\}$.   

\noindent\textbf{Hypothesis Updating} In conventional PatchMatch multi-view stereo methods~\cite{Zheng2014PatchMatch,Schonberger2016Pixelwise,Xu2018Multi}, in order to determine the best hypothesis from the candidate hypothesis set, the following multi-view aggregated matching cost is defined by photometric consistency to measure the multi-view similarity, 
\begin{equation}\label{eq:1}
c_{\text{photo}}(\theta_{i})=\frac{\sum_{j} w_{j}\cdot m_{i,j}}{\sum_{j} w_{j}},
\end{equation}
where $m_{i,j}$ is the matching cost between the reference patch and its corresponding source patch observed on source image $I_{j}$ via $\theta_{i}$ and $w_{j}$ is the view selection weight of $I_{j}$. As the photometric consistency is unreliable in low-textured areas, these methods always fail in these areas.

\begin{figure}[t]
	\centering
	\subfigure[]{
		\includegraphics[width=0.4\linewidth]{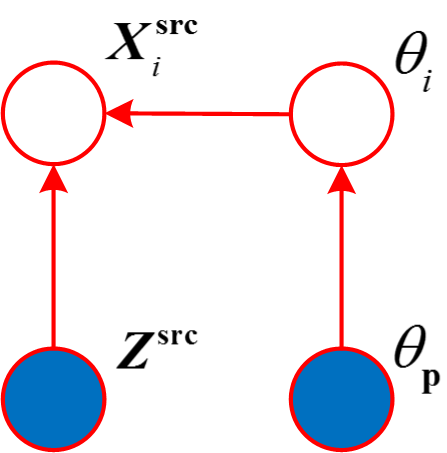}}
	\subfigure[]{
		\includegraphics[width=0.42\linewidth]{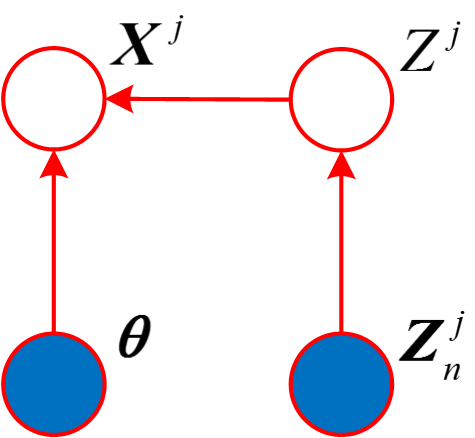}}
	\caption{(a) Graphical model of planar prior assistance. Given priori plane hypothesis $\theta_\text{p}$, the observation $\boldsymbol{X}_{i}^\text{src}$ on source images and the visibility information $\boldsymbol{Z}^\text{src}$, the optimal hypothesis $\theta^{*}$ is inferred. (b) Graphical model of view selection. At each iteration, given candidate hypotheses $\boldsymbol{\theta}$, the observation $\boldsymbol{X}^{j}$ corresponding to the hypotheses on source image $I_{j}$, and the visibility of neighboring pixels on source image $I_j$ is $\boldsymbol{Z}_{n}^{j}$, the visibility of pixel $l$ on source image $I_{j}$, $Z^{j}$, is inferred.}
	\label{fig:PELAS}
\end{figure}

In contrast, given the planar priors as described before, we leverage a probabilistic graphical model to derive our novel multi-view aggregated matching cost. To construct the graphical model, we define the patch on pixel $l$ of $I_\text{ref}$ as $X^{\text{ref}}$. Also, the patches observed on all source images via $\theta_{i}$ are $\boldsymbol{X}_{i}^{\text{src}}$, the visibility information of all source images is assumed to be $\boldsymbol{Z}^{\text{src}}$ and the planar prior at pixel $l$ is $\theta_{\text{p}}=[d_{\text{p}},\boldsymbol{n}_{\text{p}}]$. Then, the graphical model of our approach is depicted in Figure~\ref{fig:PELAS}a. The joint probability is 
\begin{equation}
P(\theta_{i},\boldsymbol{X}_{i}^{\text{src}},\boldsymbol{Z}^{\text{src}},\theta_{\text{p}}){\propto}P(\boldsymbol{X}_{i}^{\text{src}}|\theta_{i},\boldsymbol{Z}^{\text{src}})P(\theta_{i}|\theta_{\text{p}}).
\end{equation}
In this way, the maximum a-posteriori estimate of the plane hypothesis $\theta^{*}$ is given by 
\begin{equation}\label{eq:3}
\theta^{*}=\displaystyle\mathop{\arg\max}P(\theta_{i}|\boldsymbol{X}_{i}^{\text{src}},\boldsymbol{Z}^{\text{src}},\theta_{\text{p}}).
\end{equation} 
The above posterior can be factorized as
\begin{equation}\label{eq:4}
P(\theta_{i}|\boldsymbol{X}_{i}^{\text{src}},\boldsymbol{Z}^{\text{src}},\theta_{\text{p}}){\propto}P(\boldsymbol{X}_{i}^{\text{src}}|\theta_{i},\boldsymbol{Z}^{\text{src}})P(\theta_{i}|\theta_{\text{p}}).
\end{equation} 

Next, we define the likelihood function as follows,
\begin{equation}\label{eq:5}
P(\boldsymbol{X}_{i}^{\text{src}}|\theta_{i},\boldsymbol{Z}^{\text{src}})=e^{-\frac{c_{\text{photo}}(\theta_{i})^2}{\alpha}}.
\end{equation}
This function encodes the photometric consistency, making the low multi-view aggregated photometric consistency cost have high probability. It encourages our whole algorithm to choose the hypothesis with lower multi-view aggregated photometric consistency cost, which is consistent with the hypothesis update criteria in the conventional PatchMatch multi-view stereo methods. However, due to the unreliability of photometric consistency in low-textured areas, this likelihood function will not reflect hypothesis changes in these areas. In this case, it is important to leverage planar priors to reflect these changes. Thus, we define the planar prior as
\begin{equation}\label{eq:6}
P(\theta_{i}|\theta_{\text{p}})=\gamma+e^{-\frac{(d_{i}-d_{\text{p}})^2}{2\lambda_\text{d}}}\cdot e^{-\frac{{\arccos^2}\boldsymbol{n}_{i}^{\top}\boldsymbol{n}_{\text{p}}}{2\lambda_{\text{n}}}},
\end{equation}
where $\lambda_\text{d}$ is the bandwidth of depth difference and $\lambda_\text{n}$ is the bandwidth of normal difference. The planar prior encourages the propagated hypotheses to be close to the planar model at pixel $l$. We substitute Equation~(\ref{eq:4})-(\ref{eq:6}) into Equation~(\ref{eq:3}) and take the negative logarithm algorithm to get the following planar prior assisted multi-view aggregated matching cost
\begin{equation}\label{eq:7}
c_\text{p-photo}(\theta_{i})=\frac{c_\text{photo}(\theta_{i})^2}{\alpha}-\log[\gamma+e^{-\frac{(d_{i}-d_{\text{p}})^2}{2\lambda_\text{d}}}\cdot e^{-\frac{{\arccos^2}\boldsymbol{n}_{i}^{\top}\boldsymbol{n}_{\text{p}}}{2\lambda_{\text{n}}}}].
\end{equation} 
Note that, the first term that encodes photometric consistency is the main component in the above equation. This means that the photometric consistency will change more obviously than the planar prior in well-textured areas. Moreover, when the photometric consistency cannot reflect hypothesis changes in low-textured areas, the planar prior will play a major role in the hypothesis updating.

As mentioned above, the photometric consistency is important to determine the depth information in well-textured areas. This can rectify the erroneous depth estimates induced by planar models in no-planar areas. According to Equation~(\ref{eq:1}), the reliability of multi-view aggregated photometric consistency depends upon the view selection weights. To calculate these weights, we design another probabilistic graphical model to make full use of the photometric consistency of different source images and the view selection information of neighboring pixels.     

Specifically, we denote that the visibility of pixel $l$ on source image $I_{j}$ is $Z^{j}$, the visibility of neighboring pixels of pixel $l$ on source image $I_{j}$ is $\boldsymbol{Z}_{n}^{j}$, the candidate hypotheses are $\boldsymbol{\theta}$, the patch on pixel $l$ in the reference image is $X^\text{ref}$, and its corresponding patches observed on source image $I_{j}$ via $\boldsymbol{\theta}$ are $\boldsymbol{X}^{j}=\{X_{i}^{j}|i=0\cdots8\}$. The graphical model is depicted in Figure~\ref{fig:PELAS}b. According to the states of neighboring pixels, 
the joint probability is 
\begin{equation}
P(\boldsymbol{X}^{j},Z^{j},\boldsymbol{\theta},\boldsymbol{Z}_{n}^{j}){\propto}P(\boldsymbol{X}^{j}|Z^{j},\boldsymbol{\theta})P(Z_{l}^{j}|\boldsymbol{Z}_{n}^{j}),
\end{equation}
where $P(\boldsymbol{X}^{j}|Z^{j},\boldsymbol{\theta})=\sum_{i=0}^{8}P(X_{i}^{j}|Z^{j},\theta_{i})$ independently models the possible source image subset for each hypothesis while $P(Z^{j}|\boldsymbol{Z}_{n}^{j})=\sum_{l'\in \mathcal{N}(l)} P(Z^{j}|Z_{l'}^{j})$ models the smoothness of the view selection of neighboring pixels. Note that, $\mathcal{N}(l)$ stands for the four neighboring pixels of pixel $l$. Specifically, we define $P(X_{i}^{j}|Z^{j},\theta_{i})$ as
\begin{equation}
P(X_{i}^{j}|Z^{j},\theta_{i})=e^{-\frac{m_{i,j}^2}{2\sigma^2}},
\end{equation}
where $\sigma$ is a constant. And, $P(Z^{j}|Z_{l'}^{j})$ is defined as
\begin{equation}
P(Z^{j}|Z_{l'}^{j})=\left\{ \begin{array}{ll}
\eta, & \textrm{if}\;Z_{l}^{j}=Z_{l'}^{j}; \\
1-\eta, & {\textrm{else.}}
\end{array} \right.
\end{equation}

According to Bayesian rule, the view selection probability of pixel $l$ on source image $I_{j}$ is 
\begin{equation}
P(Z^{j}|\boldsymbol{X}^{j},\boldsymbol{\theta},\boldsymbol{Z}_{n}^{j}){\propto}P(\boldsymbol{X}^{j}|Z^{j},\boldsymbol{\theta})P(Z^{j}|\boldsymbol{Z}_{n}^{j}).
\end{equation}
Based on the above view selection probabilities, we employ the Monte-Carlo sampling~\cite{Bishop2006PRML} to define the weight for each source image $I_{j}$ as $w_{j}$.

\noindent\textbf{Refinement} After each hypothesis updating, we refine the current selected hypothesis $\theta_{\text{c}}=[d_\text{c},\boldsymbol{n}_\text{c}]$ by generating extra candidate hypothesis set. Following \cite{Schonberger2016Pixelwise,Xu2018Multi}, this candidate hypothesis set is defined as  
\begin{equation}\label{eq:12}
\{[d_\text{p},\boldsymbol{n}_\text{c}],[d_\text{r},\boldsymbol{n}_\text{c}],[d_\text{c},\boldsymbol{n}_\text{p}],[d_\text{c},\boldsymbol{n}_\text{r}],[d_\text{r},\boldsymbol{n}_\text{r}],[d_\text{p},\boldsymbol{n}_\text{p}]\},
\end{equation}
where $d_\text{p}$ and $\boldsymbol{n}_\text{p}$ are the randomly perturbed depth and normal with respect to $\theta_\text{c}$, $d_\text{r}$ and $\boldsymbol{n}_\text{r}$ are randomly generated depth and normal. If the cost of a new hypothesis is less than that of the current hypothesis, we will set it as the current hypothesis. The above hypothesis sampling and propagation, hypothesis updating and refinement are performed iteratively to produce the depth map for the reference image.   

\subsection{Geometric Consistency}

In the previous Section, we consider both piecewise planar priors and photometric consistency to get better depth estimation in low-textured and well-textured areas. However, there still exist some errors. This attributes to some unreliable planar priors caused by some intractable erroneous sparse correspondences. To tackle these errors, we resort to multi-view geometric consistency~\cite{Schonberger2016Pixelwise,Xu2018Multi}, 
which is defined as
\begin{equation}\label{eq:13}
c_\text{geo}(\theta_{i})=\frac{\sum_{j}w_{j}\cdot (m_{i,j}+\lambda_\text{geo}\cdot \displaystyle\mathop{\min}(\Delta e_{j}(\theta_{i}),\tau_\text{geo}))}{\sum_{j}w_{j}},
\end{equation} 
where $\lambda_\text{geo}$ is a geometric consistency regularizer, $\Delta{e}_{j}(\theta_{i})$ is the reprojection error between $I_\text{ref}$ and $I_{j}$ induced by $\theta_{i}$, and $\tau_\text{geo}$ is a truncation threshold to robustify the reprojection error against occlusions.

\begin{algorithm}[t]
	\caption{Planar Prior Assisted PatchMatch MVS}   
	\label{alg:Overview}
	{\bf Input:} multi-view images with their camera parameters \\
	{\bf Output:} hypothesis maps 
	\begin{algorithmic}[1] 
		\FOR{each image}
		\STATE set reference image and source images
		\STATE randomly initialize a hypothesis map
		\FOR{iteration $i=1$ to $T_\text{photo}$}
		\STATE hypothesis sampling and propagation
		\STATE update the hypothesis map via Equation~(\ref{eq:1})
		\STATE refinement via Equation~(\ref{eq:12})
		\ENDFOR
		\ENDFOR
		\STATE sparse credible correspondences selection
		\STATE triangulation and generate planar models
		\FOR{each image}
		\STATE set reference image and source images
		\STATE randomly initialize a hypothesis map
		\FOR{iteration $i=1$ to $T_\text{p-photo}$}
		\STATE hypothesis sampling and propagation
		\STATE update the hypothesis map via Equation~(\ref{eq:7})
		\STATE refinement via Equation~(\ref{eq:12})
		\ENDFOR
		\ENDFOR
		\STATE use the hypothesis maps obtained above as extra input
		\FOR{each image}
		\STATE set reference image and source images
		\STATE use the previous obtained hypothesis map for the reference image as initialization
		\FOR{iteration $i=1$ to $T_\text{geo}$} 
		\STATE hypothesis sampling and propagation
		\STATE update the hypothesis map via Equation~(\ref{eq:13})
		\STATE refinement via via Equation~(\ref{eq:12})
		\ENDFOR 
		\ENDFOR
	\end{algorithmic} 
\end{algorithm}

\subsection{The Algorithm}

The overall pipeline of our algorithm is summarized in Algorithm~\ref{alg:Overview}. From step 1 to step 9, we generate initial depth maps via conventional multi-view aggregated photometric consistency cost. Then, in step 10 and step 11, we select credible correspondences and triangulate them to generate planar models. From step 12 to step 20, we generate plane-awareness depth maps by our proposed planar prior assisted multi-view aggregated matching cost. Then, these depth maps are used as additional input in step 21. We further optimize these depth maps via geometric consistency from step 22 to step 30. To make each PatchMatch MVS process converge, $T_\text{photo}$, $T_\text{p-photo}$ and $T_\text{geoo}$ are set to 3, 3, 2, respectively. 

\subsection{Fusion} 

The depth maps estimated for individual images always contain noise and outliers. We follow the conventional PatchMatch pipeline~\cite{Schonberger2016Pixelwise,Xu2018Multi} to use a fusion step to produce the final point cloud. Each image is treated as the reference image in turn and its depth estimates are unprojected to the world coordinate to obtain 3D points. These 3D points are further projected to neighboring images to calculate their projected depths, normals and image coordinates. According to the estimated depths and normals in projected image coordinates, a consistent estimate is determined if its relative depth difference is lower than $0.01$, normal difference is lower than $10^\circ$, and reprojection error is less than $2$ pixels. The estimate that has two consistent neighboring estimates are kept and their unprojected 3D points are averaged to produce the final 3D point. 

\begin{table*}[t]
	\caption{Percentage of pixels with absolute errors below $2cm$ and $10cm$ on the high-resolution multi-view training datasets of ETH3D benchmark (in $\%$). ACMP$\backslash$G means ACMP without geometric consistency. The related values are from \cite{Xu2018Multi}. The best results are marked in bold.}
	\centering
	\footnotesize
	\begin{tabular}{p{0.6cm}<{\centering}p{1.36cm}<{\centering}p{0.5cm}<{\centering}p{0.5cm}<{\centering}p{0.5cm}<{\centering}p{0.5cm}<{\centering}p{0.45cm}<{\centering}p{0.5cm}<{\centering}p{0.5cm}<{\centering}p{0.5cm}<{\centering}p{0.7cm}<{\centering}p{0.5cm}<{\centering}p{0.5cm}<{\centering}p{0.6cm}<{\centering}p{0.7cm}<{\centering}p{0.7cm}<{\centering}}
		\toprule 
		\multirow{2}{*}{error} & \multirow{2}{*}{method} & \multirow{2}{*}{Ave.} & \multicolumn{7}{c}{indoor} & \multicolumn{6}{c}{outdoor} \\ 
		\cmidrule(lr){4-10} \cmidrule(lr){11-16} 
		& & & deli. & kick. & offi. & pipes & relief & relief. & terrai. & courty. & elec. & faca. & mead. & playgr. & terrace \\
		\midrule 
		\multirow{6}{*}{$2 cm$} & {COLMAP} & 65.0 & 69.7 & {43.5} & 26.3 & 41.1 & 86.3 & 85.8 & 57.6 & {82.6} & 71.0 & {74.2} & 54.6 & 70.9 & 80.8 \\
		& {openMVS} & 55.2 & 63.9 & 36.9 & 29.3 & 31.8 & 80.1 & 81.5 & 57.9 & 64.3 & 55.6 & 55.4 & 29.8 & 57.9 & 73.6 \\
		& {ACMH} & 68.5 & {73.3} & 42.7 & {32.3} & {53.6} & {89.1} & {90.3} & {71.4} & 79.9 & {74.8} & 68.5 & {57.1} & {75.3} & {82.0} \\
		& {ACMM} & 80.5 & {77.7} & {\bf 66.7} & {51.2} & {76.5} & {\bf 96.0} & {95.7} & {85.4} & {84.4} & {86.8} & {74.5} & {\bf 77.1} & {\bf 84.3} & {89.7} \\
		& {ACMP$\backslash$G} & 72.9 & 76.9 & 47.4 & 48.6 & 65.3 & 91.3 & 92.6 & 79.1 & 78.8 & 79.2 & 68.9 & 59.5 & 73.7 & 85.9 \\
		& {ACMP} & {\bf 81.9} & {\bf 81.9} & 62.0 & {\bf 65.6} & {\bf 78.6} & 94.8 & {\bf 95.8} & {\bf 88.4} & {\bf 84.5} & {\bf 88.7} & {\bf 76.6} & 76.8 & 80.6 & {\bf 90.7} \\
		\midrule 
		\multirow{6}{*}{$10 cm$} & {COLMAP} & 73.7 & 80.6 & 51.4 & 34.2 & 47.8 & 89.6 & 89.3 & 63.5 & 93.4 & 77.4 & {90.9} & 70.1 & 81.0 & 89.1 \\
		& {openMVS} & 66.5 & 79.1 & 45.1 & 38.2 & 42.1 & 84.1 & 86.0 & 67.1 & 79.1 & 64.8 & 77.4 & 48.4 & 68.7 & 83.8 \\
		& {ACMH} & 79.1 & {84.2} & {51.9} & {41.8} & {61.7} & {92.3} & {94.1} & {77.8} & {93.7} & {83.4} & 90.8 & {78.6} & {86.9} & {91.5} \\
		& {ACMM} & {\bf 90.7} & {93.0} & {\bf 80.0} & {64.8} & {83.9} & {\bf 98.2} & {\bf 98.4} & {90.4} & {\bf 97.3} & {\bf 94.7} & {\bf 93.4} & {\bf 91.7} & {\bf 95.1} & {\bf 98.0} \\
		& {ACMP$\backslash$G} & 85.1 & 90.7 & 61.3 & 66.6 & 74.8 & 94.5 & 96.6 & 86.1 & 93.7 & 88.6 & 90.7 & 80.0 & 86.8 & 95.3 \\
		& {ACMP} & 90.6 & {\bf 95.4} & 72.4 & {\bf 78.0} & {\bf 84.2} & 96.8 & 98.2 & {\bf 92.4} & 96.1 & 94.4 & 93.2 & 87.5 & 91.5 & 97.9 \\
		\bottomrule 
	\end{tabular}
	\label{tab:depthmap}
\end{table*}

\begin{figure*}
	\centering	
	\includegraphics[width=0.119\textwidth]{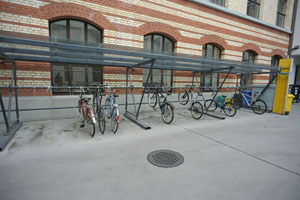} 
	\includegraphics[width=0.119\textwidth]{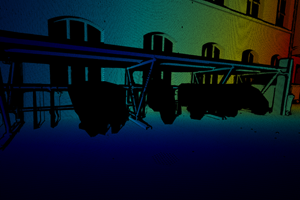} 
	\includegraphics[width=0.119\textwidth]{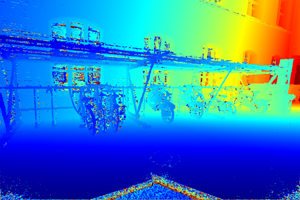}
	\includegraphics[width=0.119\textwidth]{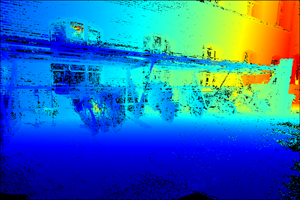}
	\includegraphics[width=0.119\textwidth]{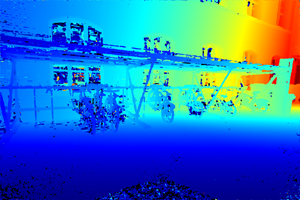}
	\includegraphics[width=0.119\textwidth]{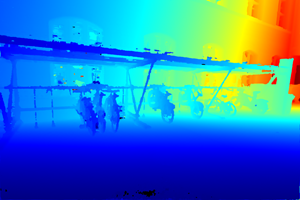}
	\includegraphics[width=0.119\textwidth]{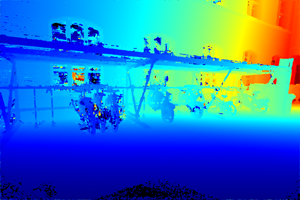}
	\includegraphics[width=0.119\textwidth]{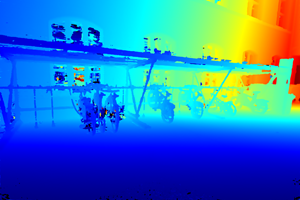} \\
	\includegraphics[width=0.119\textwidth]{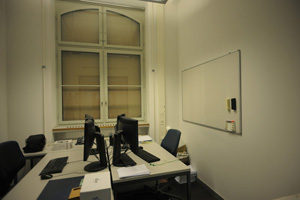} 
	\includegraphics[width=0.119\textwidth]{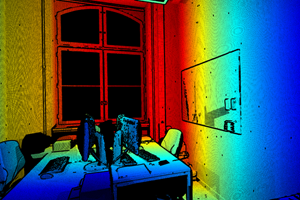}
	\includegraphics[width=0.119\textwidth]{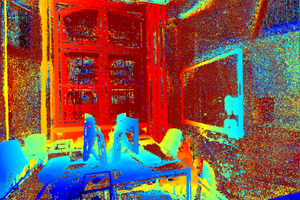}
	\includegraphics[width=0.119\textwidth]{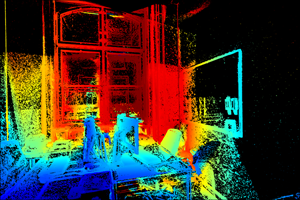}
	\includegraphics[width=0.119\textwidth]{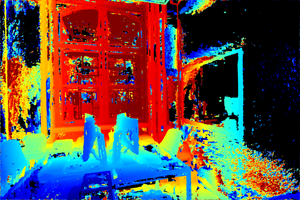}
	\includegraphics[width=0.119\textwidth]{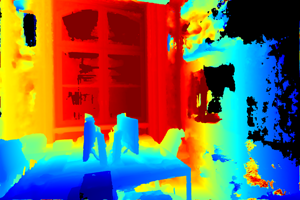}
	\includegraphics[width=0.119\textwidth]{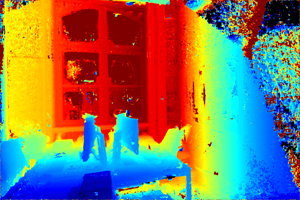}
	\includegraphics[width=0.119\textwidth]{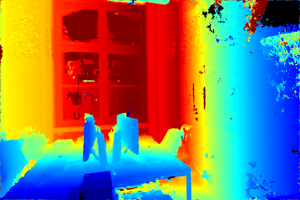} \\
	\includegraphics[width=0.119\textwidth]{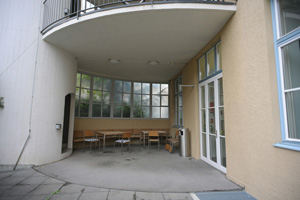} 
	\includegraphics[width=0.119\textwidth]{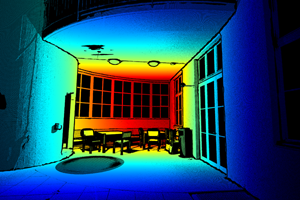}
	\includegraphics[width=0.119\textwidth]{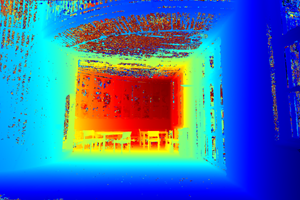}
	\includegraphics[width=0.119\textwidth]{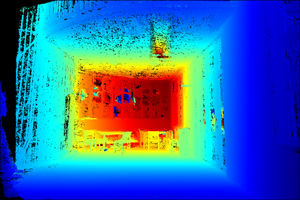}
	\includegraphics[width=0.119\textwidth]{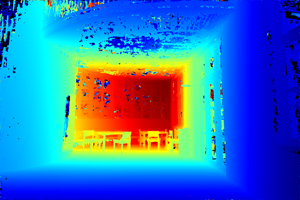}
	\includegraphics[width=0.119\textwidth]{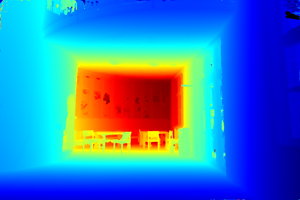}
	\includegraphics[width=0.119\textwidth]{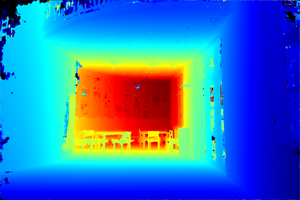}
	\includegraphics[width=0.119\textwidth]{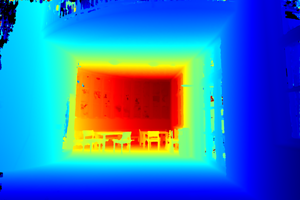} \\
	\begin{tabular}{>{\centering}p{0.1\textwidth}>{\centering}p{0.1\textwidth}>{\centering}p{0.1\textwidth}>{\centering}p{0.1\textwidth}>{\centering}p{0.1\textwidth}>{\centering}p{0.1\textwidth}>{\centering}p{0.1\textwidth}>{\centering}p{0.1\textwidth}}
		Image & GT & COLMAP & openMVS & ACMH & ACMM & ACMP$\backslash$G & ACMP
	\end{tabular}
	\caption{Qualitative depth map map comparisons of different methods on some high-resolution multi-view training datasets (courty., offi. and terrace) of ETH3D benchmark. Black pixels in GT mean no ground truth data.}
	\label{fig:dme}
\end{figure*}

\section{Experiments} 

\noindent{\bf Datasets} We evaluate the effectiveness of our method on high-resolution multi-view stereo dataset of ETH3D benchmark~\cite{Schops2017Multi}. This dataset contains images at a resolution of $6048\times4032$ with calibration. Following \cite{Schonberger2016Pixelwise,Xu2018Multi}, we resize this imagery to no more than $3200$ pixels for each dimension while keeping the original aspect ratio. The dataset is further split into training datasets and test datasets. Besides ground truth point clouds, ground truth depth maps are also provided for training datasets. Thus, we first evaluate the depth estimation on training datasets. As for test datasets, we submit our reconstructed point clouds to the benchmark's website~\cite{Schops2017ETH3D} to evaluate them.

\noindent{\bf Evaluation Metrics} In depth map evaluation, we calculate the percentage of pixels with an absolute depth error less than $2cm$ and $10cm$ from ground truth. For point cloud evaluation, we assess reconstructed point clouds in terms of accuracy, completeness and $F_{1}$ score. 

\noindent{\bf Parameter Settings} Our methods are implemented in C++ with CUDA and executed on a machine with two Intel E5-2630 CPUs and two GTX Titan X GPUs. $\{\epsilon,\alpha,\gamma, \lambda_\text{n}, \sigma,\eta,\lambda_\text{geo},\tau_\text{geo}\}=\{0.1,0.18,0.5,5^\circ,0.3,0.9,0.1,5.0\}$. Besides, $\lambda_\text{d}$ is adaptively set to one sixty-fourth of the depth interval of every reference image. We conduct geometric consistency twice as \cite{Xu2018Multi} does.

\noindent{\bf Depth Map Evaluation} We compare our method with some state-of-the-art PatchMatch multi-view stereo methods in depth map evaluation, including COLMAP~\cite{Schonberger2016Pixelwise}, openMVS~\cite{openMVS}, ACMH~\cite{Xu2018Multi} and ACMM~\cite{Xu2018Multi}. These methods are directly operated on the original resolution images except for ACMM with multi-scale scheme. We denote our method as ACMP because our planar prior assisted model is based on adaptive checkerboard sampling and propagation.

\begin{figure*}[t]
	\centering
	\includegraphics[width=0.16\textwidth]{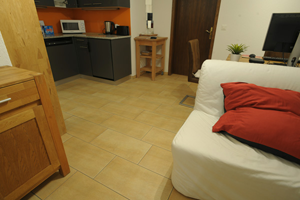} 
	\includegraphics[width=0.16\textwidth]{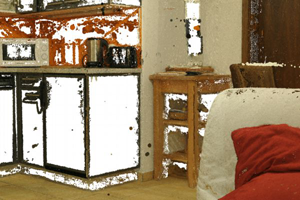} 
	\includegraphics[width=0.16\textwidth]{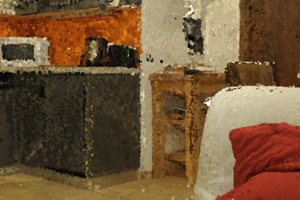}
	\includegraphics[width=0.16\textwidth]{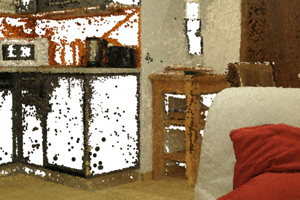}
	\includegraphics[width=0.16\textwidth]{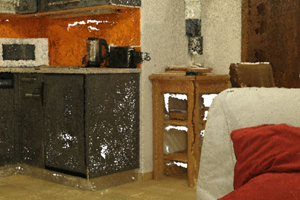}
	\includegraphics[width=0.16\textwidth]{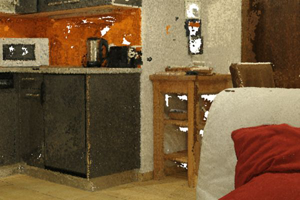}
	\includegraphics[width=0.16\textwidth]{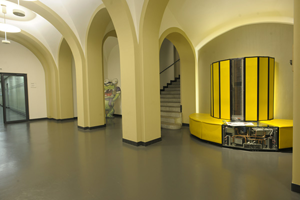} 
	\includegraphics[width=0.16\textwidth]{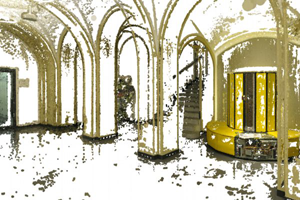} 
	\includegraphics[width=0.16\textwidth]{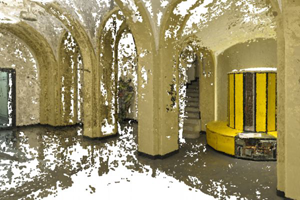}
	\includegraphics[width=0.16\textwidth]{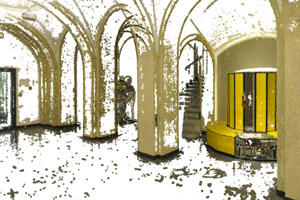}
	\includegraphics[width=0.16\textwidth]{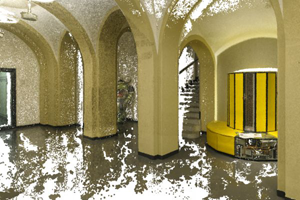}
	\includegraphics[width=0.16\textwidth]{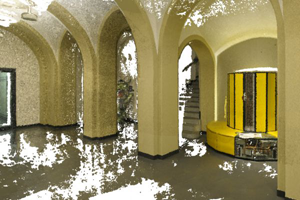}
    \begin{tabular}{p{0.14\textwidth}<{\centering}p{0.14\textwidth}<{\centering}p{0.14\textwidth}<{\centering}p{0.14\textwidth}<{\centering}p{0.14\textwidth}<{\centering}p{0.14\textwidth}<{\centering}}
	Images & COLMAP & openMVS & ACMH & ACMM & ACMP
    \end{tabular}
	\caption{Qualitative point cloud comparisons of different MVS methods on living room and old computer of ETH3D benchmark. These 3D models are reported by the ETH3D benchmark evaluation server~\cite{Schops2017ETH3D}.}
	\label{fig:pce}
\end{figure*} 

We list comparison results on $13$ high resolution multi-view training datasets of ETH3D benchmark in Table~\ref{tab:depthmap}. In order to validate the effectiveness of our novel multi-view aggregated matching cost, we remove the geometric consistency from our method and denote this method as ACMP$\backslash$G. We first compare ACMP$\backslash$G with COLMAP, openMVS and ACMH. As can be seen, ACMP$\backslash$G is much better than these methods. From the qualitative results in Figure~\ref{fig:dme}, we observe that ACMP$\backslash$G can estimate depth information of low-textured areas well. Moreover, ACMP$\backslash$G can also tackle the depth estimation in non-planar regions because our planar prior assisted multi-view aggregated matching cost simultaneously considers the photometric consistency. 

As ACMP$\backslash$G does not consider the geometric consistency, its estimated depth maps contain some noise caused by inappropriate planar models. Therefore, ACMP combines ACMP$\backslash$G with geometric consistency to handle this problem. This makes our method competitive with ACMM. Note that, ACMM uses multi-scale geometric consistency to tackle the depth estimation in low-textured areas. However, due to its limited scales and lost image information at its coarsest scale, ACMM sometimes cannot obtain good estimation for low-textured areas at the coarsest scale. This can be reflected by the depth estimation of offi. dataset in Figure~\ref{fig:dme}. Differently, as ACMP adaptively captures the discrimination of different sizes according to triangular primitives, ACMP performs much better than ACMM in offi. dataset.

\begin{table}[!htb]
	\caption{Accuracy, completeness and $F_{1}$ score (in $\%$) comparisons of reconstructed point clouds on the high-resolution multi-view test datasets of ETH3D benchmark at evaluation threshold $2cm$. The related values are from \cite{Schops2017ETH3D}.}
	\centering
	\footnotesize 
	\begin{tabular}{ccccc}
		\toprule
		& method & Accuracy & Completeness & $F_{1}$ \\
		\midrule
		\multirow{5}{*}{indoor} & COLMAP & {\bf 91.95} & 59.65 & 70.41 \\
		& openMVS & 82.00 & {\bf 75.92} & 78.33 \\
		& ACMH & 91.14 & 64.81 & 73.93 \\
		& ACMM & 90.99 & 72.73 & 79.84 \\
		& ACMP & 90.60 & 74.23 & {\bf 80.57} \\
		\midrule
		\multirow{5}{*}{outdoor} & COLMAP &  {\bf 92.04} & 72.98 & 80.81 \\
		& openMVS & 81.93 & {\bf 86.41} & 84.09 \\
		& ACMH & 83.96 & 80.03 & 81.77 \\
		& ACMM & 89.63 & 79.17 & 83.58 \\
		& ACMP & 90.35 & 79.62 & {\bf 84.36} \\
		\midrule
		\multirow{5}{*}{all} & COLMAP & {\bf 91.97} & 62.98 & 73.01 \\
		& openMVS & 81.98 & {\bf 78.54} & 79.77 \\
		& ACMH & 89.34 & 68.62 & 75.89 \\
		& ACMM & 90.65 & 74.34 & 80.78  \\
		& ACMP & 90.54 & 75.58 & {\bf 81.51} \\
		\bottomrule 
	\end{tabular}
\label{tab:pointcloud}
\end{table}

\noindent{\bf Point Cloud Evaluation} The quantitative results of reconstructed point clouds are listed in Table~\ref{tab:pointcloud} and quantitative results are shown in Figure~\ref{fig:pce}. In the case of $2cm$, ACMP achieves the best $F_{1}$ score among all methods due to our better estimated depth maps. As for the completeness of 3D models, openMVS produces the best performance as it employs a different fusion scheme with a relaxed number of consistent views for a pixel, which leads to noisy point clouds with low accuracy and high completeness. As illustrated in Figure~\ref{fig:pce}, the 3D models of openMVS are not photo-realistic. Differently, our method can achieve a better trade-off between accuracy and completeness. This makes our estimated 3D models be applicable in the actual environment.

\begin{table}
	\caption{Running time of different stages of our method for an image of size $3200\times2130$ pixels on a single GPU.}
	\centering
	\footnotesize
	\begin{tabular}{ccc}
		\toprule
		Stage & Time (s) & Ratio ($\%$) \\
		\midrule
		Sparse Correspondences Generation & 6.40 & 30.9 \\
		Planar Model Construction & 1.08 & 5.2 \\
		Planar Prior Assistance & 5.43 & 26.2 \\
		Geometric Consistency & 7.79 & 37.6 \\
		\midrule
		Total Time & 20.7 & - \\
		\bottomrule
	\end{tabular}
\label{tab:Time}
\end{table}

\noindent{\bf Runtime Analysis} We run our method on a single GPU and record the running time of each stage in Table~\ref{tab:Time}. As can be seen, the planar model construction occupies very little runtime. As for sparse correspondences generation, planar prior assistance and geometric consistency, their running time is very close. This is because these stages all employ the same pipeline of PatchMatch multi-view stereo. Therefore, with very little computational cost, our method without geometric consistency can achieve much better reconstruction results than other PatchMatch multi-view stereo methods that are implemented on the original image resolution.

\section{Conclusion}

In this work, we propose a planar prior assisted PatchMatch multi-view stereo framework to help the depth estimation in low-textured areas. To tackle depth ambiguities caused by unreliable photometric consistency, we leverage sparse credible correspondences to build planar models. These models can reflect the depth ranges of low-textured areas well but are still biased, especially for non-planar regions. Therefore, we embed these planar models into PatchMatch multi-view stereo by utilizing a probabilistic graphical model. This derives a novel multi-view aggregated matching cost, which jointly consists of photometric consistency and planar priors. This makes our method suited for both planar and non-planar regions. Experiments on ETH3D benchmark demonstrate the effectiveness of our methods by yielding state-of-the-art performance. We note that the performance of our method is comparable to ACMM. Since ACMM combines ACMH with the multi-scale geometric consistency, we think its performance gain mainly comes from the multi-scale geometric consistency framework. As a separate module, we have demonstrated that our method is much better than ACMH in our experiments. Thus, we may combine our method with the multi-scale geometric consistency to improve the reconstruction performance in the future work.

\noindent\textbf{Acknowledgements} This work was supported by the National Natural Science Foundation of China under Grants 61772213 and 91748204.  

\bibliographystyle{aaai}
\bibliography{refs}

\end{document}